\documentclass{article}
\usepackage{arxiv}

\usepackage{amsmath,amssymb,amsfonts}
\usepackage{algorithmic}
\usepackage{algorithm}
\usepackage{array}
\usepackage[caption=false,font=normalsize,labelfont=sf,textfont=sf]{subfig}
\usepackage{textcomp}
\usepackage{stfloats}
\usepackage{url}
\usepackage{verbatim}
\usepackage{graphicx}
\usepackage{cite}
\usepackage{adjustbox}
\usepackage{booktabs}
\usepackage{algorithmic}
\usepackage{multirow}
\usepackage{arydshln}
\usepackage{appendix}

\usepackage{tikz}
\usepackage{calc} 
\usetikzlibrary{calc} 

\usepackage[
    bookmarksopen,
    bookmarksdepth=2,
    breaklinks=true
]{hyperref}

\definecolor{ZoomRed}{RGB}{255,0,0}

\newcommand{\imgwithzoom}[7]{%
\begin{tikzpicture}

    \node[anchor=south west, inner sep=0] (MainImg) at (0,0) {\includegraphics[#1]{#2}};
    
    \path (MainImg.south west); \pgfgetlastxy{\Mx}{\My}
    \path (MainImg.north east); \pgfgetlastxy{\Mw}{\Mh}
    
    \pgfmathsetmacro{\SourceBoxFrac}{#5} 
    \pgfmathsetmacro{\InsetBoxFrac}{#6}  
    
    \pgfmathsetmacro{\SourceSize}{\SourceBoxFrac * \Mw}
    \pgfmathsetmacro{\InsetSize}{\InsetBoxFrac * \Mw}
    
    \pgfmathsetmacro{\Mag}{\InsetSize / \SourceSize}
    
    \pgfmathsetmacro{\CenterRelX}{#3}
    \pgfmathsetmacro{\CenterRelY}{#4}
    
    \pgfmathsetmacro{\Cx}{\CenterRelX * \Mw}
    \pgfmathsetmacro{\Cy}{\CenterRelY * \Mh}
    
    \coordinate (SrcCenter) at (\Cx pt, \Cy pt);
    \coordinate (SrcSW) at (\Cx pt - 0.5*\SourceSize pt, \Cy pt - 0.5*\SourceSize pt);
    \coordinate (SrcNE) at (\Cx pt + 0.5*\SourceSize pt, \Cy pt + 0.5*\SourceSize pt);
    
    \def\BorderPad{0.01*\Mw} 
    \coordinate (InsetSW) at (\Mw - \BorderPad - \InsetSize pt, \BorderPad);
    \coordinate (InsetNE) at (\Mw - \BorderPad, \BorderPad + \InsetSize pt);
    \coordinate (InsetCenter) at ($ (InsetSW)!0.5!(InsetNE) $);

    \begin{scope}
        \clip (InsetSW) rectangle (InsetNE);
        
        \fill[white] (InsetSW) rectangle (InsetNE);
        
        \pgfmathsetmacro{\ShiftX}{\Cx * \Mag}
        \pgfmathsetmacro{\ShiftY}{\Cy * \Mag}
        
        \coordinate (PlacementPoint) at ($ (InsetCenter) + (-\ShiftX pt, -\ShiftY pt) $);
        
        \node[anchor=south west, inner sep=0, scale=\Mag] at (PlacementPoint) 
             {\includegraphics[#1]{#2}};
    \end{scope}

    \ifnum#7=1
        \draw[ZoomRed, line width=1pt] (SrcSW) rectangle (SrcNE);
    \fi
    
    \draw[ZoomRed, line width=1pt] (InsetSW) rectangle (InsetNE);

\end{tikzpicture}%
}

\makeatletter
\def\adl@drawiv#1#2#3{%
        \hskip.5\tabcolsep
        \xleaders#3{#2.5\@tempdimb #1{1}#2.5\@tempdimb}%
                #2\z@ plus1fil minus1fil\relax
        \hskip.5\tabcolsep}
\newcommand{\cdashlinelr}[1]{%
  \noalign{\vskip\aboverulesep
           \global\let\@dashdrawstore\adl@draw
           \global\let\adl@draw\adl@drawiv}
  \cdashline{#1}
  \noalign{\global\let\adl@draw\@dashdrawstore
           \vskip\belowrulesep}}
\makeatother

\newcommand{\x}{\boldsymbol{x}}
\newcommand{\y}{\boldsymbol{y}}

\newcommand{\e}{\boldsymbol{e}}
\newcommand{\K}{\boldsymbol{K}}
\renewcommand{\v}{\boldsymbol{v}}
\newcommand{\vtheta}{\boldsymbol{v}_\Theta}

\begin{document}

\title{Physics-Guided Flow Matching for CT Image Reconstruction}

\author{ Davide Evangelista \\
  Department of Computer Science and Engineering\\
  University of Bologna\\
  Bologna, 40126, Italy \\
  \texttt{davide.evangelista5@unibo.it}}

\maketitle

\begin{abstract}
Deep generative models have recently emerged as powerful priors for solving ill-posed inverse problems in computed tomography (CT), with diffusion-based approaches achieving state-of-the-art reconstruction performance. However, diffusion models typically rely on stochastic sampling procedures, long inference trajectories, and carefully tuned noise schedules, which can limit computational efficiency and numerical stability, especially at high spatial resolutions. In this work, we investigate Flow Matching as an alternative generative prior for CT reconstruction.
We train a high-resolution Rectified Flow Matching model on $256 \times 256$ chest images from the Mayo Clinic Low-Dose CT dataset. To mitigate overfitting and limited anatomical variability, we employ a two-stage training strategy consisting of an initial phase with strong, anatomically informed data augmentation, followed by a fine-tuning phase with reduced or no augmentation to refine structural fidelity. The resulting model is capable of generating high-quality and anatomically coherent CT-like images, serving as a strong learned prior.
We then evaluate multiple reconstruction methods specifically designed for Flow Matching models, including Plug-and-Play Flow, FlowDPS, Flower, and Flow-Priors (ICTM), and compare them against state-of-the-art diffusion-based reconstruction algorithms such as DDRM, DPS, and DiffPIR. Experimental results across several CT inverse problem settings show that Flow Matching-based approaches consistently outperform diffusion-based methods in terms of PSNR, SSIM, and perceptual quality, while requiring fewer sampling steps.
Finally, we publicly release the trained Flow Matching model and accompanying code to facilitate reproducibility and future research. Overall, this work demonstrates that Flow Matching provides a stable, efficient, and effective alternative to diffusion models for high-resolution CT image reconstruction.
\end{abstract}


\section{Introduction}
\label{sec:introduction}
Computed Tomography (CT) \cite{sidky2008image,buzug2011computed,zeng2001image} is a fundamental application of modern medical imaging, providing high-resolution visualization of internal anatomical structures for diagnosis, treatment planning, and longitudinal monitoring. Despite its clinical importance, CT reconstruction remains a fundamentally ill-posed inverse problem, particularly in low-dose, sparse-view, or limited-angle acquisition settings, where the non-injectivity of the acquisition operator gives rise to a reconstruction problem with infinitely many admissible solutions, thereby violating Hadamard’s second condition of well-posedness \cite{hadamard1902sur}. These scenarios motivate the development of reconstruction algorithms capable of recovering high-fidelity images from incomplete or noisy measurements while preserving anatomical plausibility.

In recent years, deep generative models have emerged as powerful data-driven priors for CT reconstruction \cite{kim2025flowdps,kawar2022denoising}. Among them, diffusion-based models \cite{ho2020denoising} have demonstrated remarkable performance across a wide range of inverse problems, including CT, by leveraging iterative stochastic sampling schemes to approximate the posterior distribution. Methods such as DDRM \cite{kawar2022denoising} and DPS \cite{chung2023diffusion} have shown that diffusion priors can substantially improve reconstruction quality compared to classical optimization-based approaches. However, diffusion models typically require long sampling chains, careful noise scheduling, and stochastic solvers, leading to high computational cost and limited flexibility at inference time. These limitations become particularly pronounced in high-resolution medical imaging, where reconstruction speed, numerical stability, and reproducibility are critical.

Flow Matching (FM) \cite{lipman2023flow} has been recently proposed as an alternative generative modeling framework that avoids explicit stochastic diffusion processes. Instead, FM learns a continuous-time velocity field that deterministically transports samples from a simple base distribution to the data distribution by solving an ordinary differential equation (ODE). This formulation offers several advantages over diffusion models, including deterministic sampling, reduced sensitivity to hyperparameter choices, and natural compatibility with higher-order numerical solvers. Rectified Flow Matching \cite{liu2022flow} further simplifies training while retaining strong generative performance. Despite these appealing properties, the application of Flow Matching to medical imaging and CT reconstruction remains largely unexplored, with existing studies mostly limited to low-resolution natural images or preliminary experiments.

In this work, we investigate Flow Matching as a generative prior for high-resolution CT reconstruction. We train a Rectified Flow Matching model on the Mayo Clinic Low-Dose CT dataset \cite{moen2021low}, consisting of $3{,}306$ acquisitions of human chests at a spatial resolution of $256 \times 256$ pixels from $10$ different patients. Training expressive generative models on medical data poses significant challenges due to limited anatomical variability and the risk of overfitting, especially at high spatial resolutions and with small-sized datasets, as is typical in medical imaging scenarios. To address this issue, we design a carefully balanced augmentation pipeline that includes elastic deformations, intensity perturbations, and geometric transformations, aimed at increasing data diversity while preserving anatomical realism. This strategy enables stable training and results in a model capable of generating high-quality CT-like images, which we use as an indicator of the expressiveness of the learned prior.

While strong augmentation improves generalization, it may also introduce subtle anatomical distortions that are undesirable in a clinical context. To mitigate this effect, we introduce a subsequent fine-tuning phase with reduced or no augmentation, allowing the model to refine structural details and improve anatomical fidelity. The resulting model exhibits improved visual coherence and stability. We publicly release the trained weights and generation code to facilitate reproducibility and future research. To the best of our knowledge, this is the first publicly available high-resolution Flow Matching model trained on CT data.

\begin{figure*}[tbhp!]
    \centering
    \includegraphics[width=0.9\linewidth]{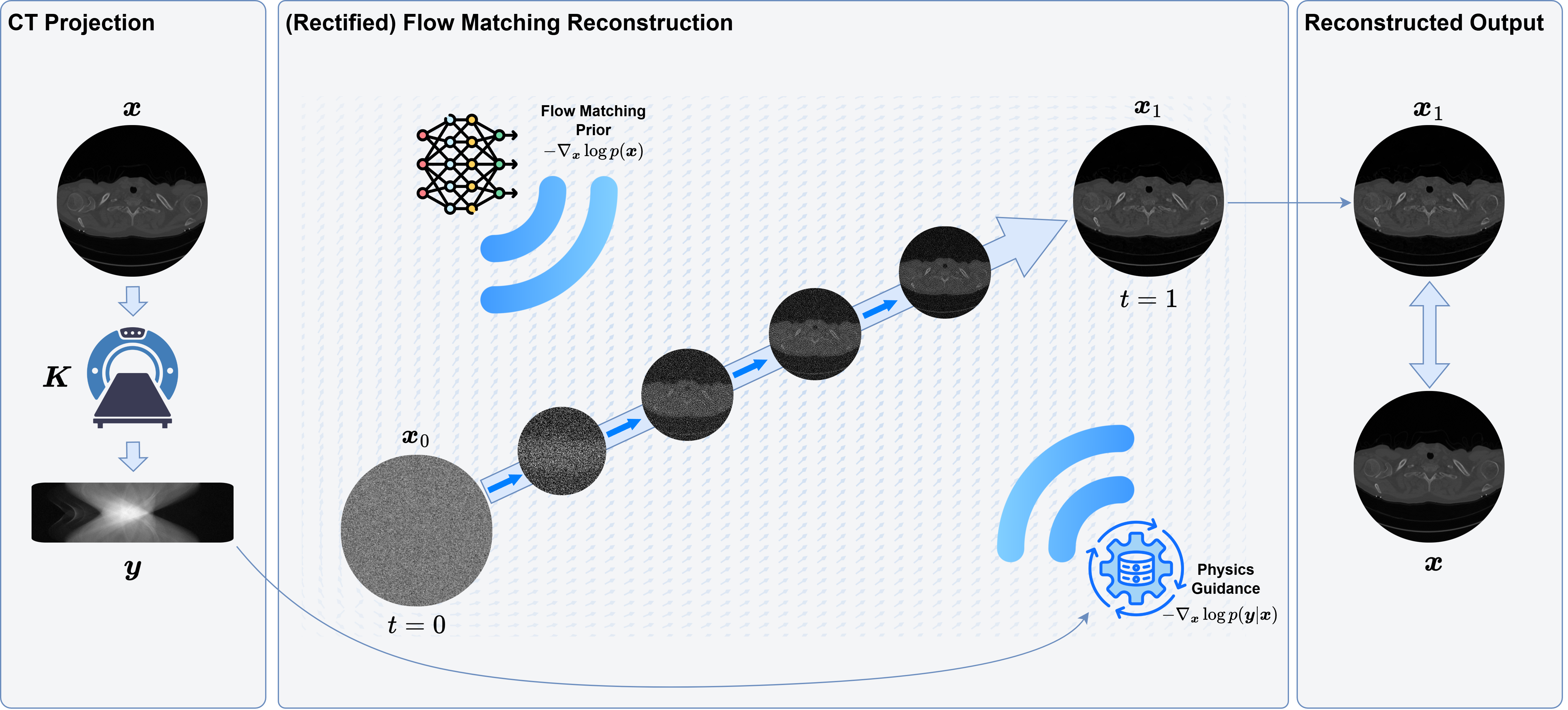}
    \caption{Graphical overview of the proposed Flow Matching model and its use as a generative prior for CT reconstruction.}
    \label{fig:graphical_abstract}
\end{figure*}

In this work, we evaluate the effectiveness of Flow Matching priors for solving sparse-view CT inverse problems. We consider several reconstruction strategies specifically developed for Flow Matching models, including Plug-and-Play Flow (PnP-Flow) \cite{martin2025pnpflow}, FlowDPS \cite{kim2025flowdps}, FLOWER \cite{pourya2025flower}, and Flow-Priors (ICTM) \cite{zhang2024flowpriors}. These methods are compared against state-of-the-art diffusion-based reconstruction algorithms such as DDRM \cite{kawar2022denoising} and DPS \cite{chung2023diffusion} under identical experimental settings. Reconstruction performance is assessed using standard quantitative metrics, including PSNR, SSIM \cite{wang2004image}, and LPIPS \cite{zhang2018unreasonable}, while generation quality is evaluated using Kernel Inception Distance (KID) \cite{binkowski2018demystifying} together with qualitative visual inspection. Our results show that Flow Matching-based methods consistently achieve superior reconstruction and generation quality, particularly in preserving fine anatomical details and reducing reconstruction artifacts, even when using a small number of integration steps. A schematic overview of how trained Flow Matching model are employed in CT reconstruction is shown in Figure~\ref{fig:graphical_abstract}.

In summary, the main contributions of this work are as follows: (i) we present a high-resolution Rectified Flow Matching model trained on CT-like medical images and demonstrate its ability to learn a strong and anatomically meaningful prior; (ii) we introduce a two-stage training strategy combining aggressive augmentation and targeted fine-tuning to balance generalization and anatomical fidelity; (iii) we provide a comprehensive comparison between Flow Matching-based and diffusion-based reconstruction methods for CT inverse problems, showing consistent improvements in reconstruction accuracy and perceptual quality; and (iv) we publicly release the trained model and code, enabling reproducible research on Flow Matching for medical image reconstruction.

The remainder of the paper is organized as follows. Section~\ref{sec:related_work} reviews related work on generative priors for inverse problems and Flow Matching models. Section~\ref{sec:method} introduces the proposed framework and reconstruction algorithms. Section~\ref{sec:experiments} presents experimental results, and Section~\ref{sec:conclusion} concludes the paper.

\section{Related Work}
\label{sec:related_work}
\noindent
CT reconstruction has been extensively studied for several decades, with classical approaches rooted in analytical and iterative methods. Analytical techniques such as filtered back-projection (FBP) \cite{zeng2001image} provide fast reconstructions under ideal sampling conditions, but suffer from severe artifacts in sparse-view, limited-angle, or noisy acquisition settings \cite{buzug2011computed}. Iterative reconstruction methods, including algebraic reconstruction techniques and statistical approaches such as the widely-used SIRT \cite{gilbert1972iterative} and SART \cite{andersen1984simultaneous} algorithms, address these limitations by explicitly modeling the acquisition process and noise statistics \cite{sidky2008image,zeng2001image}. However, due to the ill-posedness of the inverse problem, these methods require carefully designed regularization terms, which often rely on handcrafted priors and may struggle to preserve fine anatomical details.

Variational formulations incorporating total variation or sparsity-based regularization have been widely adopted to improve reconstruction quality in ill-posed settings \cite{sidky2008image,chambolle2010introduction}. While effective in reducing noise and artifacts, these approaches typically involve a trade-off between noise suppression and loss of structural detail, and their performance strongly depends on parameter tuning and prior assumptions.

The advent of deep learning has significantly advanced CT reconstruction by enabling data-driven prior modeling. Early learning-based approaches focused on direct inversion, either by post-processing FBP \cite{morotti2021green} or model-based \cite{evangelista2023rising} reconstructions, or by learning end-to-end mappings from measurements to images. While these methods demonstrated impressive empirical performance, they often lack robustness to changes in acquisition geometry and noise levels.
More recent approaches integrate deep neural networks into iterative reconstruction frameworks, combining data consistency with learned priors \cite{bianchi2025data}. Plug-and-play \cite{cascarano2022plug} and unrolled \cite{xiang2021fista} optimization methods have been widely explored in this context, offering improved interpretability and generalization. Nevertheless, most of these approaches rely on discriminative models that implicitly encode priors through supervised training, rather than explicitly modeling the underlying image distribution.

Diffusion-based generative models \cite{ho2020denoising} have recently emerged as powerful tools for modeling complex image distributions and have been successfully applied to inverse problems, including CT reconstruction. By learning a stochastic denoising process, diffusion models enable approximate posterior sampling through iterative refinement.
Several methods have adapted diffusion models to inverse problems by incorporating measurement consistency into the sampling process. DDRM \cite{kawar2022denoising} formulates posterior sampling under linear degradations with known noise statistics, while DPS \cite{chung2023diffusion} generalizes this idea by modifying the diffusion dynamics using gradient-based data consistency terms. These approaches have demonstrated state-of-the-art performance across a range of inverse problems, including sparse-view CT.
Despite their success, diffusion-based methods typically require long sampling trajectories, stochastic solvers, and carefully tuned noise schedules. These characteristics result in high computational cost and can limit numerical stability and reproducibility, particularly in high-resolution medical imaging scenarios.

Flow Matching \cite{lipman2023flow} has recently been proposed as an alternative generative modeling framework that replaces stochastic diffusion processes with deterministic continuous-time dynamics. By learning a velocity field that transports samples from a base distribution to the data distribution through an ordinary differential equation, Flow Matching enables deterministic sampling and avoids explicit noise scheduling. Rectified Flow Matching \cite{liu2022flow} further simplifies training while maintaining strong generative performance.
ODE-based generative models offer several appealing properties for inverse problems, including compatibility with higher-order numerical solvers and improved stability. However, their application to medical imaging has so far been limited. Most existing studies focus on low-resolution natural images or synthetic datasets, and only a few recent works have begun to explore Flow Matching in the context of inverse problems \cite{martin2025pnpflow,kim2025flowdps,pourya2025flower,zhang2024flowpriors}.

Among these, Plug-and-Play Flow (PnP-Flow) \cite{martin2025pnpflow} integrates Flow Matching priors into iterative reconstruction schemes by alternating between data-consistency updates and flow-based prior steps. FlowDPS \cite{kim2025flowdps} adapts diffusion posterior sampling ideas to Flow Matching by modifying the velocity field to incorporate measurement information. FLOWER \cite{pourya2025flower} proposes an alternative formulation that directly embeds data consistency into the flow dynamics, while Flow-Priors (ICTM) \cite{zhang2024flowpriors} formulate reconstruction as a constrained optimization problem guided by continuous-time flows.

While these methods demonstrate the potential of Flow Matching for inverse problems, existing evaluations are primarily limited to low-resolution images or non-medical datasets. Moreover, a systematic comparison between Flow Matching-based and diffusion-based reconstruction methods in high-resolution CT settings remains largely unexplored.

In contrast to prior studies, this work provides a comprehensive evaluation of Flow Matching models for high-resolution CT image reconstruction. By training a Rectified Flow Matching prior on $256 \times 256$ CT images and systematically comparing multiple Flow Matching-based reconstruction strategies against state-of-the-art diffusion-based methods, we aim to assess the practical advantages and limitations of Flow Matching in a clinically relevant setting.

\section{Methodology}
\label{sec:method}
\noindent
In this section, we describe the proposed framework for learning Flow Matching priors from CT data and for using them to solve CT inverse problems. We begin by introducing the inverse problem formulation and the generative prior model, then describe the training strategy adopted to learn high-resolution CT priors. Finally, we present the reconstruction methods based on Flow Matching and discuss numerical integration schemes.

\subsection{Inverse Problem Formulation}

We consider the problem of reconstructing a CT image $\x \in \mathbb{R}^{H \times W}$ from indirect measurements $\y \in \mathbb{R}^{M}$ obtained through a forward (linear) operator $\K$, namely
\begin{align}\label{eq:forward_problem}
\y = \K \x + \e,
\end{align}
where $\e \sim \mathcal{N}(\boldsymbol{0}, \sigma^2 \boldsymbol{I})$ denotes measurement noise. The operator $\K$ models the CT acquisition process and may correspond to sparse-view, limited-angle, or low-dose geometries. Due to the ill-posedness of the inverse problem, stable reconstruction requires incorporating prior information on the unknown image $\x$. In a Bayesian framework, this amounts to modeling the posterior distribution $p(\x | \y)$, which is proportional to the product of the likelihood $p(\y | \x)$ and a prior distribution $p(\x)$. A common approach is to compute the Maximum A Posteriori (MAP) estimate, defined as
\begin{align*}
\x^* = \arg\min_{\x \in \mathcal{X}} \; - \log p(\y | \x) - \log p(\x),
\end{align*}
where $\mathcal{X}$ denotes the set of feasible solutions and $\x^*$ is the reconstructed image. While the likelihood term $- \log p(\y | \x)$ admits a closed-form expression derived from the forward model in \eqref{eq:forward_problem}, the prior term $- \log p(\x)$ is considerably more challenging, as it requires modeling the complex distribution of medical images.
In this work, we model the prior distribution of CT images using a Flow Matching generative model learned from data.

\subsection{Flow Matching Prior}

Flow Matching aims to learn a continuous-time transformation that maps samples from a simple base distribution to samples from the data distribution. Let $\x_0 \sim p_0$ denote a sample from a base distribution (typically an isotropic Gaussian), and let $\x_1 \sim p_{\text{data}}$ denote a sample from the target CT image distribution. Flow Matching defines a time-dependent state $\x_t \in \mathbb{R}^{H \times W}$ for $t \in [0,1]$, governed by the ordinary differential equation
\begin{equation}
\frac{d \x_t}{dt} = \v^*(\x_t, t),
\end{equation}
where $\v^*(\cdot, t) : \mathbb{R}^{H \times W} \to \mathbb{R}^{H \times W}$ denotes the \emph{true time-dependent velocity field} associated with the Flow Matching process. In practice, this velocity field is approximated during training by a time-conditioned neural network $\vtheta(\x_t, t)$.

In this work, we adopt the Rectified Flow Matching formulation, in which the interpolation path between $\x_0$ and $\x_1$ is defined as
\begin{equation}
\x_t = (1 - t)\x_0 + t \x_1,
\end{equation}
which yields a constant target velocity field
\begin{equation}
\v^*(\x_t, t) := \frac{d \x_t}{dt} = \x_1 - \x_0.
\end{equation}
The network parameters $\Theta$ are thus learned by minimizing the Flow Matching loss
\begin{equation}
\mathcal{L}_{\text{FM}}(\Theta)
=
\mathbb{E}_{\x_0, \x_1, t}
\left[
||
\vtheta(\x_t, t) - (\x_1 - \x_0)
||_2^2
\right].
\end{equation}

In our implementation, $\vtheta(\x_t, t)$ is parameterized by a time-conditioned \texttt{ResidualAttentionUNet} architecture \cite{oktay2018attention}, consisting of four resolution levels with $(64, 128, 256, 256)$ channels, where self-attention is applied at the two deepest resolution levels. Time conditioning is achieved using a sinusoidal positional embedding \cite{vaswani2017attention}, followed by a sequence of upsampling layers to match the spatial resolution of $\x_t$. The resulting embedding is concatenated to the input along the channel dimension. Group Normalization is applied at the beginning of each block to stabilize training and mitigate the effects of the limited batch size, which is set to $4$ due to computational constraints.

Once trained, the generative model defines a deterministic mapping from $\x_0 \sim p_0$ to $\x_1 \sim p_{\text{data}}$ by numerically integrating the ODE from $t=0$ to $t=1$. In this work, we consider a first-order explicit Euler method which, given a time step $\Delta t > 0$, updates $\x_t$ as
\begin{equation}\label{eq:euler_step}
\x_{t+\Delta t} = \x_t + \Delta t \vtheta(\x_t, t),
\end{equation}

However, the deterministic nature of Flow Matching allows the use of higher-order solvers such as the Runge-Kutta methods, resulting in improved sample quality and reconstruction stability, particularly in the low steps scenario. Exploring the behavior of Flow Matching models applied to the solution of inverse problems with a higher-order ODE solver is left as future work.

\subsection{Training Strategy for High-Resolution CT Data}

Training expressive generative priors on medical images is challenging due to limited anatomical variability and the risk of overfitting, particularly at high spatial resolutions. To address this issue, we adopt a two-stage training strategy.

In the first stage, the Flow Matching model is trained using an extensive data augmentation pipeline designed to increase sample diversity while preserving anatomical plausibility. The augmentations include elastic deformations, random rotations, and intensity jittering, applied on-the-fly during training. The goal of this stage is to encourage the model to learn robust global anatomical features rather than memorizing individual samples. More specifically, we apply random rotations in the range $[-90^\circ, 90^\circ]$, horizontal flips with probability $p=0.5$, and elastic deformations with scaling factor $\alpha_{\text{AUG}} = 120$ and smoothing factor $\sigma_{\text{AUG}} = 6$, implemented using the \texttt{ElasticTransform} operator from the \texttt{Albumentations} package. During this stage, the model is trained for $500$ epochs using the \texttt{AdamW} optimizer with default parameters and a fixed learning rate of $10^{-4}$, requiring approximately $50$ hours on local hardware. The batch size is set to $4$ due to memory constraints.

\begin{figure}[!t] 
    \centering
    \begin{tabular}{cc}
        \includegraphics[width=0.45\linewidth]{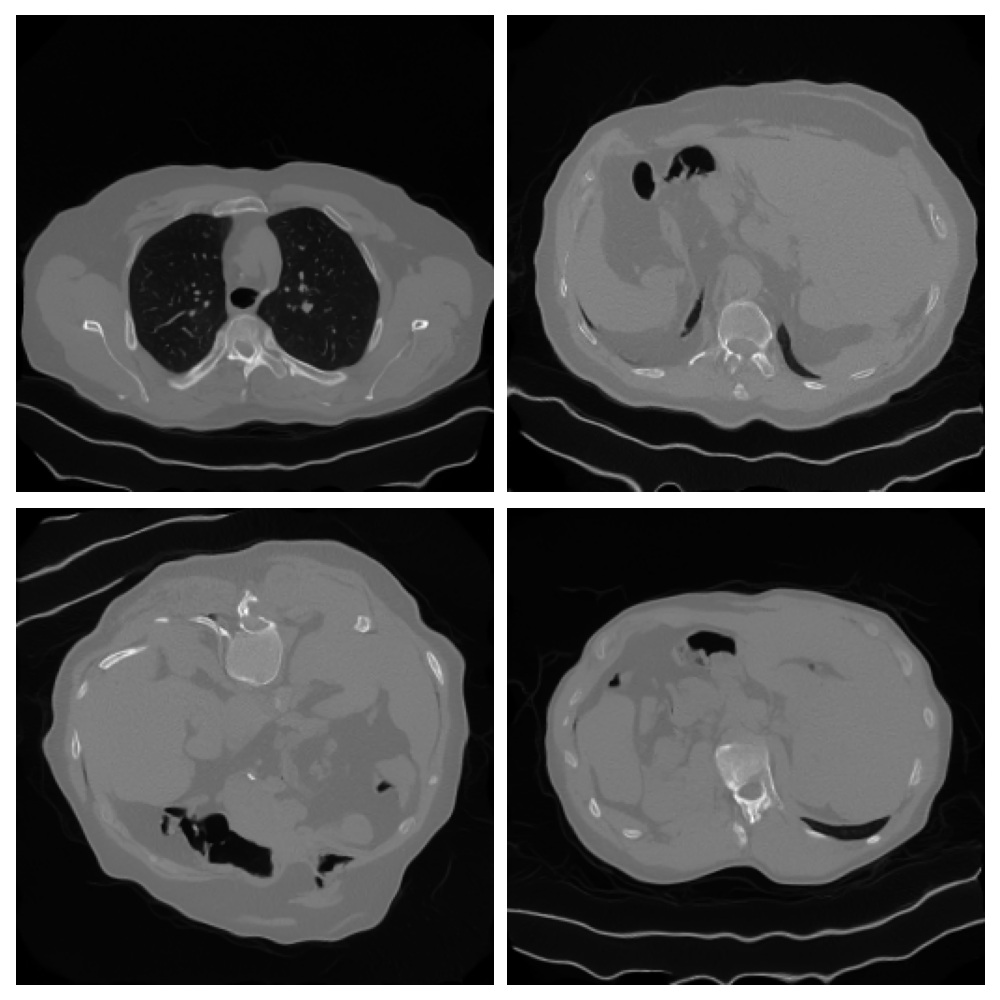} & 
        \includegraphics[width=0.45\linewidth]{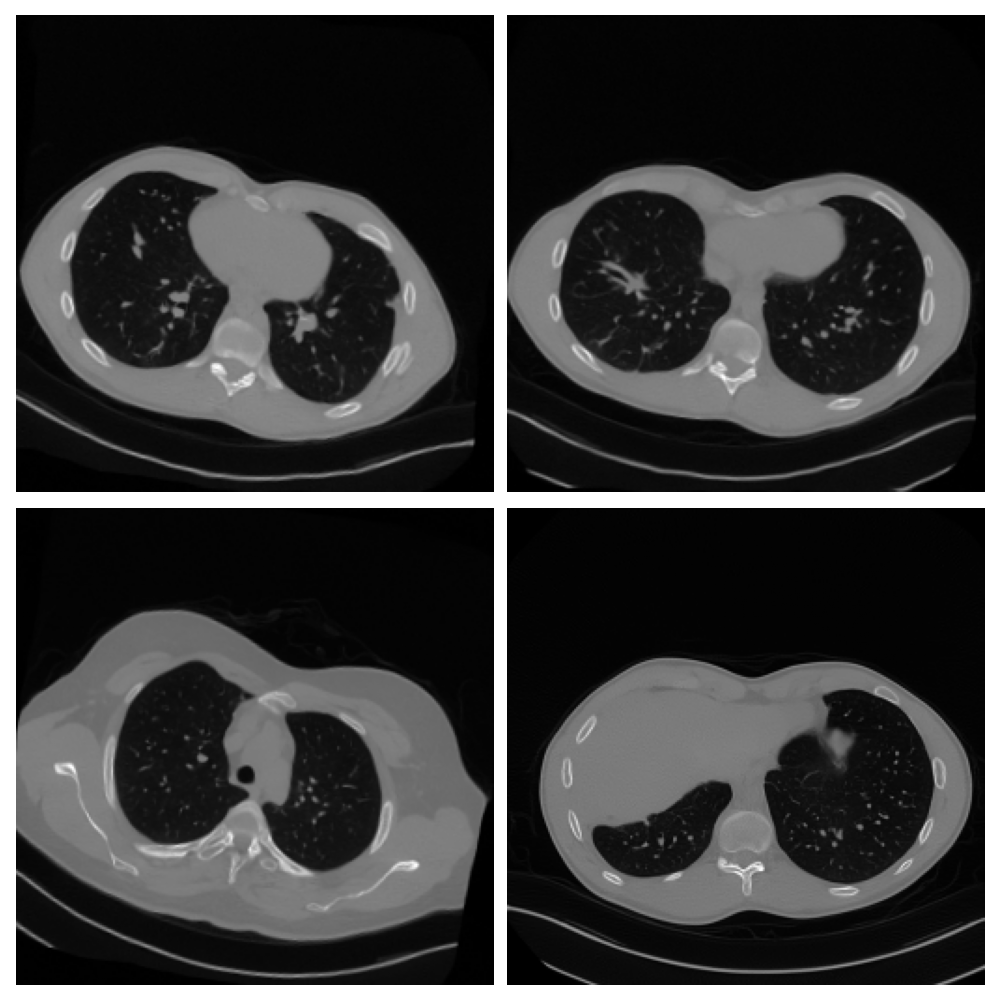} \\
        (a) & (b)
    \end{tabular}
    \caption{Samples generated by MedicalFM starting from random noise $\x_0 \sim \mathcal{N}(\boldsymbol{0}, \boldsymbol{I})$: (a) after training with strong augmentation; (b) after fine-tuning with reduced augmentation.}
    \label{fig:generation}
\end{figure}

While aggressive augmentation improves generalization, it may also introduce subtle anatomical distortions that are undesirable in a medical imaging context, as illustrated in Fig.~\ref{fig:generation}(a). To mitigate this effect, we perform a second training stage in which the model is fine-tuned using reduced augmentation, i.e. no rotations and elastic deformations with $\alpha_{\text{AUG}} = 15$ and $\sigma_{\text{AUG}} = 4$. This refinement phase allows the learned velocity field to better align with anatomically faithful CT structures while retaining the generalization benefits acquired during the initial training.

Since the model is already capable of generating realistic images after the first stage, this fine-tuning phase requires only a limited number of epochs (specifically, $100$ epochs), using the same optimizer with a reduced learning rate of $5 \cdot 10^{-5}$. As shown in Fig.~\ref{fig:generation}(b), the resulting model exhibits reduced augmentation artifacts and improved visual quality compared to the initial training stage.

\subsection{Flow Matching for CT Reconstruction}

Given a trained Flow Matching prior, we use it to solve CT inverse problems by incorporating measurement consistency into the generative dynamics. In this setting, the unknown image $\x$ to be reconstructed is identified with the initial condition of the flow, i.e., $\x_0 = \x$, and reconstruction is performed by evolving $\x_t$ from $t=0$ toward $t=1$ under a modified velocity field.

We investigate several Flow Matching-based reconstruction strategies, including Plug-and-Play Flow (PnP-Flow), FlowDPS, FLOWER, and Flow-Priors (ICTM). Although differing in their formulation, these approaches share a common principle: reconstruction is carried out by evolving a continuous-time state $\x_t$ under a velocity field that balances the learned prior dynamics with data-consistency forces derived from the forward operator $\K$.

Assuming additive Gaussian noise in the measurement model \eqref{eq:forward_problem}, the negative log-likelihood is given by
\begin{equation}
- \log p(\y | \x) = \frac{1}{2\sigma^2} || \K \x - \y ||_2^2.
\end{equation}
Consequently, the gradient of the negative log-posterior distribution at any time $t$ can be written as
\begin{equation}
\begin{split}
- \nabla_{\x_t} \log p(\x_t | \y)
&= - \nabla_{\x_t} \log p(\x_t) - \nabla_{\x_t} \log p(\y | \x_t) \\
&\approx \vtheta(\x_t, t) + \nu_t \K^{T} \left( \K \x_t - \y \right),
\end{split}
\end{equation}
where $\vtheta(\x_t, t)$ serves as an approximation of $-\nabla_{\x_t} \log p(\x_t)$ provided by the Flow Matching model, and $\nu_t$ denotes a time-dependent scaling parameter that absorbs all multiplicative constants. The specific form of $\nu_t$ and the numerical integration scheme depend on the reconstruction method under consideration. In particular, starting from a pre-trained Flow Matching model, a \emph{denoised} estimate at time $t$ can be extracted via a flow matching version of Tweedie's formula \cite{kim2025flowdps}. Under the assumption of a linear interpolation path, $\x_t = (1-t)\x_0 + t\x_1$, this relation admits the explicit form
\begin{align}\label{eq:tweedie_linear}
\begin{split}
&\hat{\x}_{0 | t} = \x_t - t\,\vtheta(\x_t,t), \\
&\hat{\x}_{1 | t} = \x_t + (1-t)\,\vtheta(\x_t,t),
\end{split}
\end{align}
which provides estimates of the clean image and the corresponding data sample, respectively. These denoised estimates constitute the key mechanism through which Flow Matching priors are coupled with measurement fidelity.

From this viewpoint, the reconstruction methods considered in this work can be interpreted as different strategies for combining a denoising map extracted from $\vtheta$ with data-consistency enforcement. FlowDPS incorporates measurement information by applying a likelihood-gradient correction to the denoised estimate $\hat{\x}_{0 \mid t}$ within each generative step, thereby embedding data consistency directly into the flow dynamics. FLOWER and ICTM instead enforce consistency through explicit projection or proximal updates in image space (often available in closed form), followed by re-entry into the generative trajectory via a controlled re-corruption step. Finally, PnP-Flow adopts a plug-and-play formulation in which the classical data-consistency gradient $\K^{T}(\K\x - \y)$ is combined with a time-dependent flow-based denoiser, replacing explicit regularization with the learned Flow Matching prior.

\section{Experimental Results}
\label{sec:experiments}
\noindent
This section evaluates Flow Matching (FM) priors for sparse-view CT reconstruction and compares them against diffusion-based baselines under matched experimental conditions. We first describe the dataset, forward model, and evaluation protocol, then discuss the practical adjustments required to make posterior sampling stable in the CT setting. We subsequently present an experiment on unconditional generation, reporting Kernel Inception Distance (KID) scores and qualitative samples for the Flow Matching model after the first and second stages of the training strategy, as well as for the diffusion model, across an increasing number of time steps. Finally, we report quantitative reconstruction results (PSNR/SSIM/LPIPS) under three angular configurations and provide a qualitative comparison. 

All the experiments are conducted on an NVIDIA RTX A4000 GPU with 16GB of V-RAM. The trained model weights and the code used for generation and reconstruction are publicly released to ensure reproducibility\footnote{Code will be provided upon acceptance.}.

\subsection{Experimental Setup}

All experiments are conducted on $256 \times 256$ chest CT slices from the Mayo Clinic Low-Dose CT dataset \cite{moen2021low}. Images are normalized to the range $[-1,1]$ for training and inference of both the Flow Matching and the diffusion models. Reconstruction quality is evaluated using PSNR and SSIM \cite{wang2004image}, which measure fidelity and structural similarity, respectively, and LPIPS \cite{zhang2018unreasonable}, which better correlates with perceptual similarity. For completeness, we also report qualitative reconstructions on representative samples in Figure~\ref{fig:qualitative_comparison}, while additional cases (together with a complete list of the hyperparameters used) are provided in the supplementary materials\footnote{Supplementary materials are available in the supporting documents tab}.

To simulate sparse-view acquisition, we consider the measurement model
\begin{equation}
\y = \K(\x_{\mathrm{phys}}) + \boldsymbol{e}, 
\qquad 
\boldsymbol{e}\sim \mathcal{N}(\boldsymbol{0},\sigma^2 \boldsymbol{I}),
\label{eq:exp_meas}
\end{equation}
with $\sigma=0.01$, where $\x_{\mathrm{phys}}\in[0,1]^{H\times W}$ denotes the physical image representation and $\K$ is the CT projection operator. We test three angular configurations, namely $60$, $90$, and $120$ projection angles, corresponding to increasingly well-conditioned reconstructions, all in the $[0, \pi]$ angular range. The operator $\K$ is implemented via the parallel-beam projector from the ASTRA toolbox, and $\K^{T}$ is implemented via backprojection. As discussed below, this choice implies that $\K^{T}$ is not an exact algebraic adjoint of $\K$ in the discrete setting, which affects algorithms that rely on adjoint-based gradients, thus requiring algorithmic corrections.

As already mentioned, we compare four FM-based reconstruction methods (FlowDPS, PnP-Flow, FLOWER, and ICTM) against diffusion-based baselines (DPS and DDRM). For diffusion models, we employ a UNet backbone with the same architecture used for Flow Matching and the same hyperparameters, adjusting the training procedure according to the typical DDPM formulation. This choice has been made to isolate the impact of the prior (diffusion vs flow) from backbone capacity. For each method, the method-specific scale parameter (denoted by $\nu_t$ in Table~\ref{tab:recon_results_by_angles}) is tuned on a validation set via grid-search for each angular configuration.

\subsection{Practical Considerations for CT Reconstruction with Generative Priors}

Applying FM- and diffusion-based posterior sampling methods to CT data requires several technical adaptations beyond the idealized assumptions typically made in inverse-problem benchmarks. A first issue is the mismatch between the image range used by the learned priors and the physical range expected by the CT operator: while the generative models operate on $\x\in[-1,1]^{H\times W}$, the forward model is defined on $\x_{\mathrm{phys}}\in[0,1]^{H\times W}$. Therefore, whenever the projection operator appears inside an iterative method (either to compute residuals or data-consistency gradients), we map the iterate to the physical range using
\begin{equation}
\mathcal{P}(\x) = \frac{\x+1}{2},
\qquad
\x_{\mathrm{phys}} = \mathcal{P}(\x),
\label{eq:normalize_phys}
\end{equation}
so that the operator is effectively evaluated as $\K(\mathcal{P}(\x))$. Conversely, corrections computed in the physical range are mapped back to the $[-1, 1]$ range, taking into account the derivative factor induced by \eqref{eq:normalize_phys} while computing gradients.

A second, more critical issue concerns the magnitude of operator-induced gradients. In CT, the normal operator $\K^{T}\K$ can have a large spectral norm, so the backprojection of the residual $\K^{T}(\K\x_t-\y)$ may become extremely large in magnitude, particularly at early iterations (when $\x_t$ is still close to noise) and under severe undersampling. This effect can cause numerical explosions when coupling the data-consistency term with a learned prior dynamics, especially in methods that balance the likelihood gradient against the learned velocity field. To mitigate this, we introduce an explicit normalization factor that scales all gradients originating from the forward model. In practice, we precompute
\begin{equation}
\gamma = \frac{1}{H \sqrt{N_\alpha}},
\label{eq:grad_norm_factor}
\end{equation}
where $H=W=256$ and $N_\alpha$ is the number of projection angles, and we multiply data-consistency gradients by $\gamma$. This simple normalization keeps the correction steps comparable across angular configurations and prevents the fidelity term from dominating the dynamics. In addition, we clamp intermediate iterates and clip gradient magnitudes when needed to avoid numerical instabilities.

Finally, the ASTRA backprojection is not guaranteed to be the exact transpose of the discrete forward projector \cite{dong2019fixing}. This discrepancy may bias gradients and degrade convergence in methods that assume access to the exact adjoint (e.g., when solving normal equations or performing projection-like corrections). To reduce sensitivity to this mismatch, we use Krylov-based least-squares corrections when required. In particular, for projection-type updates as in ICTM or DDRM, we perform a small number of iterations of CGLS on the normal equations, approximately solving
\begin{equation}
(\K^{T}\K) \x = \K^{T}\y
\label{eq:cgls_update}
\end{equation}
and thus computing an approximation of the pseudo-inverse of $\K$. Empirically, these corrections improve stability compared to naive backprojection-based steps.

\begin{figure}
    \centering
    \includegraphics[width=\linewidth]{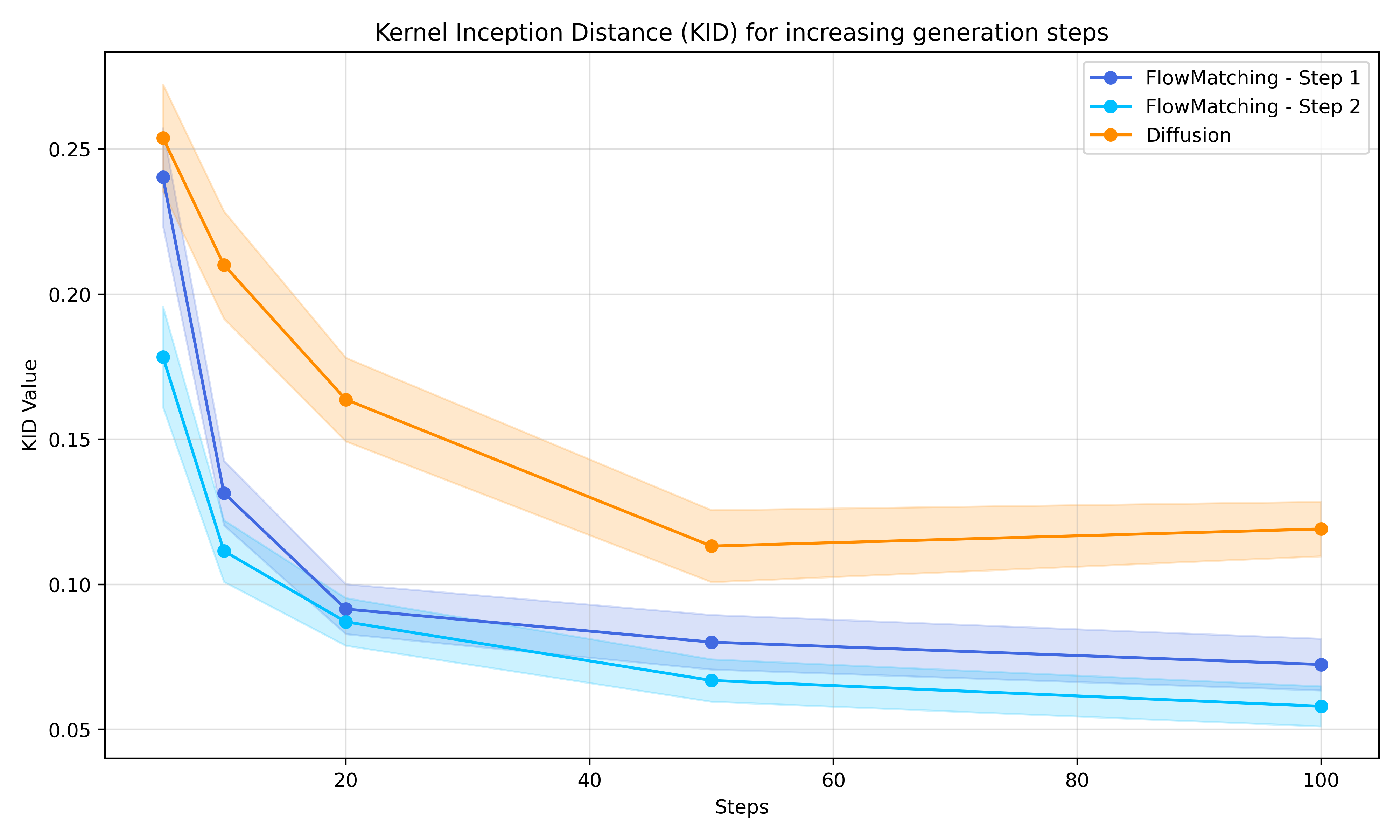}
    \caption{Kernel Inception Distance (KID) for unconditional CT image generation at increasing numbers of generation steps. We report mean and standard deviation across runs for Flow Matching after the first training stage (strong augmentation), after the second fine-tuning stage, and for a diffusion-based baseline evaluated under the same number of steps.}
    \label{fig:kid_plot}
\end{figure}

\subsection{Unconditional Generation}

Before assessing inverse-problem performance, we perform a sanity check to verify that the Flow Matching prior learns a meaningful high-resolution CT image distribution and that the proposed two-stage training strategy effectively improves sample fidelity. We generate unconditional samples by integrating the learned generative ODE from Gaussian noise up to $t=1$ using the explicit Euler method as in Equation \eqref{eq:euler_step}, considering an increasing number of integration steps $N \in \{5, 10, 20, 50, 100\}$, where $N = \frac{1}{\Delta t}$. Generation quality is evaluated using the Kernel Inception Distance (KID) \cite{binkowski2018demystifying} with 350 new images, never seen during training, of Mayo's dataset as ground-truth reference. We report KID after (i) the first training stage employing strong data augmentation and (ii) the subsequent fine-tuning stage with reduced or no augmentation. For reference, we also report KID values obtained by a diffusion-based generative model evaluated under the same number of sampling steps.

As shown in Figure~\ref{fig:kid_plot}, Flow Matching consistently achieves lower KID values than the diffusion baseline across all step counts, with a particularly pronounced gap in the low-step regime. This highlights the ability of Flow Matching to generate high-quality samples with a small number of integration steps, whereas diffusion models exhibit a stronger dependence on long sampling chains to reach comparable fidelity.

Quantitatively, the two-stage training strategy yields a substantial improvement in generation quality: KID decreases of $10\%$ to $20\%$ from first to second stage, consistently across all step counts. This confirms that the fine-tuning phase effectively mitigates artifacts introduced by aggressive augmentation, while preserving the generalization benefits learned during the first stage. Qualitatively, samples produced by the fine-tuned model, shown in Figure~\ref{fig:generation} exhibit improved anatomical coherence, sharper structures, and more realistic texture, whereas the model trained only with strong augmentation occasionally produces distortions such as unrealistic deformations or inconsistent anatomical rotations, coherent with the applied augmentations.

Overall, this experiment demonstrates both the effectiveness of Flow Matching as a high-resolution generative prior for CT images, particularly in low-step regimes, and the importance of the proposed two-stage training strategy for achieving high-fidelity and anatomically plausible samples.

\begin{table}[t]
\centering
\caption{Reconstruction results (PSNR $\uparrow$, SSIM $\uparrow$, LPIPS $\downarrow$) for different numbers of projection angles.}
\label{tab:recon_results_by_angles}
\begin{tabular}{c lcccc}
&\textbf{Method} & $\nu_t$ & \textbf{PSNR} & \textbf{SSIM} & \textbf{LPIPS} \\
\midrule
\multirow{7}{*}{\rotatebox{90}{\textbf{Angles = $60$}}}
&DPS  & $0.01$ & $28.51$ & $0.8128$ & $0.3530$ \\
&DDRM & --  & $35.30$ & $0.8934$ & $0.2768$ \\
\cdashlinelr{2-6}
&FlowDPS  & $5.0$ & $\mathbf{36.19}$ & $\mathbf{0.9162}$ & $\mathbf{0.1759}$ \\
&PnP-Flow & $0.4$ & $34.18$ & $0.8819$ & $0.2590$ \\
&FLOWER   & $50.0$ & $34.47$ & $0.8382$ & $0.2724$ \\
&ICTM     & $0.9$ & $34.51$ & $0.8598$ & $0.2878$ \\
\midrule
\midrule
\multirow{7}{*}{\rotatebox{90}{\textbf{Angles = $90$}}}
&DPS  & $0.015$ & $30.75$ & $0.8453$ & $0.3247$ \\
&DDRM  & -- & $35.89$ & $0.9080$ & $0.2545$ \\
\cdashlinelr{2-6}
&FlowDPS  & $5.0$ & $\mathbf{36.56}$ & $\mathbf{0.9230}$ & $\mathbf{0.1653}$ \\
&PnP-Flow & $0.4$ & $34.84$ & $0.8918$ & $0.2392$ \\
&FLOWER   & $50.0$ & $36.41$ & $0.9080$ & $0.2439$ \\
&ICTM     & $0.9$ & $34.65$ & $0.8694$ & $0.2672$ \\
\midrule
\midrule
\multirow{7}{*}{\rotatebox{90}{\textbf{Angles = $120$}}}
&DPS  & $0.02$ & $32.76$ & $0.8686$ & $0.2984$ \\
&DDRM  & -- & $36.21$ & $0.9200$ & $0.2449$ \\
\cdashlinelr{2-6}
&FlowDPS  & $1.0$ & $\mathbf{36.64}$ & $\mathbf{0.9260}$ & $\mathbf{0.1576}$ \\
&PnP-Flow & $0.3$ & $35.47$ & $0.9065$ & $0.2119$ \\
&FLOWER   & $25.0$ & $35.92$ & $0.9078$ & $0.1861$ \\
&ICTM     & $0.8$ & $35.41$ & $0.8968$ & $0.2090$ \\
\bottomrule
\end{tabular}
\end{table}

\begin{figure*}[tbhp]
    \centering
    \setlength{\tabcolsep}{2pt}
    \resizebox{\linewidth}{!}{
    \begin{tabular}{cc cccccc}
    & \Large{True} & \Large{DPS} & \Large{DDRM} & \Large{FlowDPS} & \Large{PnP-Flow} & \Large{FLOWER} & \Large{ICTM} \\
    \rotatebox{90}{\hspace{15px}\Large\textbf{Angles = $60$}}&
    \imgwithzoom{width=0.2\linewidth}{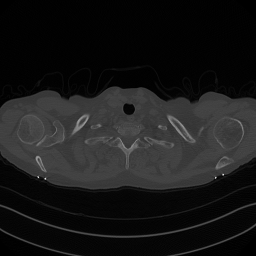}{0.5}{0.4}{0.2}{0.3}{1}&
    \imgwithzoom{width=0.2\linewidth}{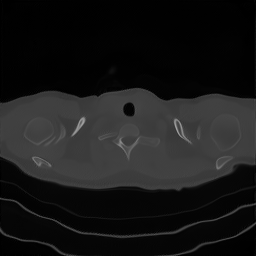}{0.5}{0.4}{0.2}{0.3}{0}&
    \imgwithzoom{width=0.2\linewidth}{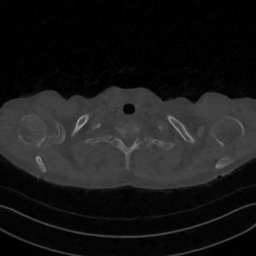}{0.5}{0.4}{0.2}{0.3}{0}&
    \imgwithzoom{width=0.2\linewidth}{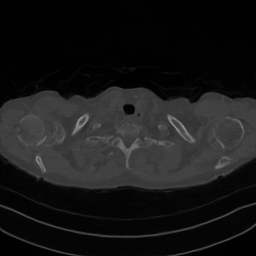}{0.5}{0.4}{0.2}{0.3}{0}&
    \imgwithzoom{width=0.2\linewidth}{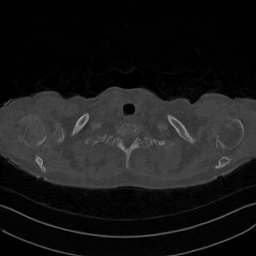}{0.5}{0.4}{0.2}{0.3}{0}&
    \imgwithzoom{width=0.2\linewidth}{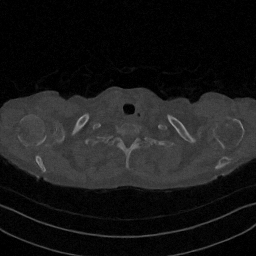}{0.5}{0.4}{0.2}{0.3}{0}&
    \imgwithzoom{width=0.2\linewidth}{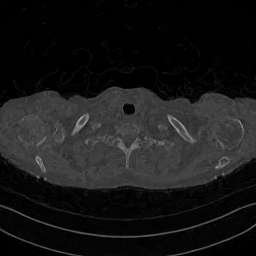}{0.5}{0.4}{0.2}{0.3}{0} \\
    
    \rotatebox{90}{\hspace{15px}\Large\textbf{Angles = $90$}}&
    \imgwithzoom{width=0.2\linewidth}{imgs/C081_35/C081_35.png}{0.5}{0.4}{0.2}{0.3}{1}&
    \imgwithzoom{width=0.2\linewidth}{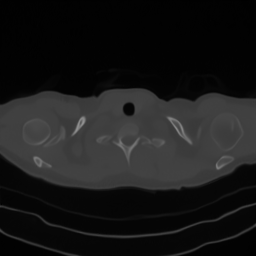}{0.5}{0.4}{0.2}{0.3}{0}&
    \imgwithzoom{width=0.2\linewidth}{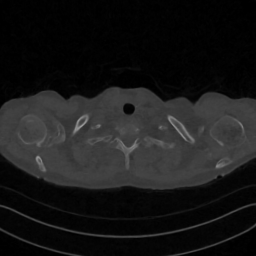}{0.5}{0.4}{0.2}{0.3}{0}&
    \imgwithzoom{width=0.2\linewidth}{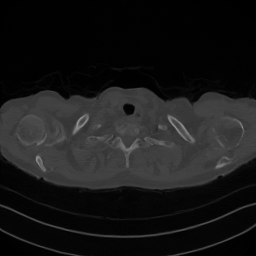}{0.5}{0.4}{0.2}{0.3}{0}&
    \imgwithzoom{width=0.2\linewidth}{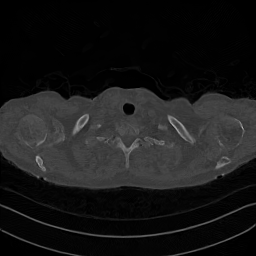}{0.5}{0.4}{0.2}{0.3}{0}&
    \imgwithzoom{width=0.2\linewidth}{imgs/C081_35/recon_flowers_st300_a60_s50.0/recon.png}{0.5}{0.4}{0.2}{0.3}{0}&
    \imgwithzoom{width=0.2\linewidth}{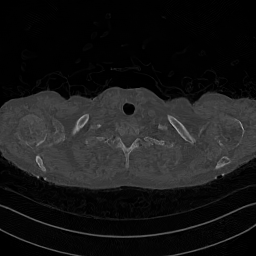}{0.5}{0.4}{0.2}{0.3}{0} \\

    \rotatebox{90}{\hspace{15px}\Large\textbf{Angles = $120$}}&
    \imgwithzoom{width=0.2\linewidth}{imgs/C081_35/C081_35.png}{0.5}{0.4}{0.2}{0.3}{1}&
    \imgwithzoom{width=0.2\linewidth}{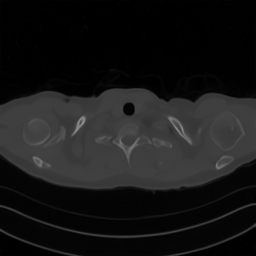}{0.5}{0.4}{0.2}{0.3}{0}&
    \imgwithzoom{width=0.2\linewidth}{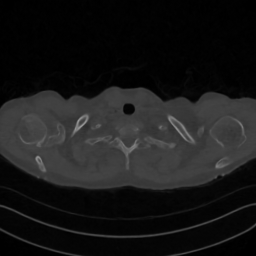}{0.5}{0.4}{0.2}{0.3}{0}&
    \imgwithzoom{width=0.2\linewidth}{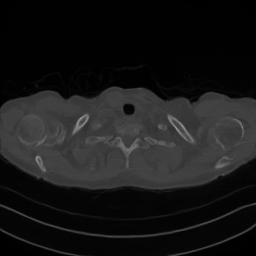}{0.5}{0.4}{0.2}{0.3}{0}&
    \imgwithzoom{width=0.2\linewidth}{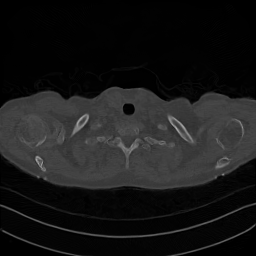}{0.5}{0.4}{0.2}{0.3}{0}&
    \imgwithzoom{width=0.2\linewidth}{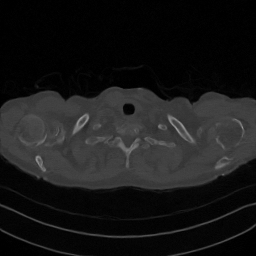}{0.5}{0.4}{0.2}{0.3}{0}&
    \imgwithzoom{width=0.2\linewidth}{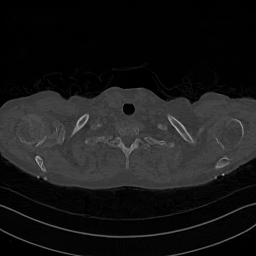}{0.5}{0.4}{0.2}{0.3}{0} \\
    \end{tabular}}
    \caption{Qualitative comparison of sparse-view CT reconstruction results. Rows correspond to different angular undersampling scenarios ($60$, $90$, and $120$ projection angles). Columns display the Ground Truth, diffusion-based baselines (DPS, DDRM), and the evaluated Flow Matching methods (FlowDPS, PnP-Flow, FLOWER, ICTM). The red insets highlight fine anatomical details, demonstrating that Flow Matching priors (particularly FlowDPS and PnP-Flow) consistently recover sharper structures and minimize artifacts compared to diffusion models, especially in the challenging $60$-angle setting.}
    \label{fig:qualitative_comparison}
\end{figure*}

\subsection{Reconstruction Results}

Table~\ref{tab:recon_results_by_angles} reports reconstruction performance using $60$, $90$, and $120$ projection angles. As expected, reconstruction quality improves as the number of angles increases, reflecting the improved conditioning of the inverse problem. Among diffusion baselines, DPS performs the worst across all configurations, highlighting the difficulty of posterior sampling under severe sparsity when the data-consistency term must be strongly enforced. DDRM provides a stronger baseline, achieving PSNR values above $35$~dB and SSIM values close to $0.90$.

Across all angular settings, Flow Matching-based approaches consistently outperform diffusion-based baselines, with particularly large gains in perceptual quality as measured by LPIPS. FlowDPS achieves the best overall performance for all three angular configurations, reaching PSNR up to $36.64$~dB, SSIM up to $0.9260$, and LPIPS as low as $0.1576$. The LPIPS gap relative to DDRM is substantial (e.g., $0.2768$ for DDRM versus $0.1759$ for FlowDPS at $60$ angles), suggesting that Flow Matching priors preserve fine anatomical texture and reduce perceptually salient artifacts more effectively than diffusion priors in this setting.

Within the family of FM methods, FLOWER and PnP-Flow are competitive in terms of PSNR and SSIM but obtain lower LPIPS when compared to FlowDPS, especially in the most challenging sparse-view regime ($60$ angles). ICTM yields consistent improvements over diffusion baselines but appears more sensitive to the choice of the data-consistency scaling parameter $\nu_t$ and to operator mismatch. This behavior is consistent with its reliance on projection-based updates and motivates our use of relaxed projections and CGLS-based corrections.

Representative reconstructions are shown in Fig.~\ref{fig:qualitative_comparison}. Diffusion-based methods are generally able to recover the global anatomical structure, but their performance degrades noticeably under severe angular sparsity. DPS produces overly smooth reconstructions across all angular settings, resulting in a pronounced loss of fine anatomical details. DDRM yields visually cleaner reconstructions and recovers most anatomical structures without introducing noticeable noise. However, as highlighted in the zoomed regions, very small structures appear slightly blurred. This effect becomes more evident at $60$ and $90$ angles, where portions of the zoomed bone structures are partially lost.

In contrast, Flow Matching-based methods consistently produce sharper reconstructions with improved preservation of anatomical details. Among all methods, FlowDPS yields the most visually plausible reconstructions across all angular configurations, effectively suppressing streaking artifacts while preserving fine structures and avoiding blur or noise amplification. Nevertheless, similarly to diffusion-based methods, FlowDPS (as well as DPS and DDRM) fails to recover some extremely small and symmetric bone structures located at the very bottom of the patient slice.

PnP-Flow and ICTM exhibit very similar qualitative behavior. Both methods generate high-quality reconstructions with well-preserved fine details and minimal blurring. However, they tend to introduce a small amount of residual noise. This observation is consistent with their slightly inferior LPIPS scores compared to FlowDPS and can be attributed to the semi-convergence behavior of the CGLS data-consistency step, even though only two CGLS iterations are employed in practice. Interestingly, unlike FlowDPS and diffusion-based methods, both PnP-Flow and ICTM are able to recover the small symmetric bone structures at the bottom of the slice. FLOWER also produces visually convincing reconstructions but similarly introduces mild noise and fails to recover these small bone structures, resulting in slightly worse qualitative performance than PnP-Flow and ICTM.

Overall, the qualitative results corroborate the quantitative findings, showing that Flow Matching-based methods substantially outperform diffusion-based approaches in terms of visual quality and anatomical coherence, particularly in severely underdetermined acquisition settings. These observations are consistent with the superior generation quality of Flow Matching priors observed in the unconditional generation experiments.

\section{Discussion}
\noindent
A key practical takeaway of these experiments is that applying generative-prior posterior sampling methods to CT requires careful numerical stabilization. The combination of (i) the mismatch between the model intensity range and the physical projection range, (ii) the amplification induced by $\K^{T}\K$, and (iii) the inexactness of backprojection as an adjoint, can lead to unstable dynamics if left unaddressed. The normalization $\mathcal{P}(\x)=(\x+1)/2$, the gradient scaling factor $\gamma$, and the use of clamping/gradient control are essential to ensure stable convergence across angular configurations, while CGLS-based corrections mitigate the impact of adjoint mismatch in projection-type updates.

Overall, the results demonstrate that Flow Matching priors provide a highly effective alternative to diffusion priors for sparse-view CT reconstruction. Across all angular configurations, FM-based methods achieve higher PSNR/SSIM and substantially lower LPIPS than diffusion-based baselines, with FlowDPS yielding the best overall performance. The unconditional generation sanity-check further supports the effectiveness of the proposed two-stage training strategy, as reflected by improved KID after fine-tuning and by enhanced anatomical fidelity in generated samples.

\section{Conclusions}
\label{sec:conclusion}
\noindent
In this work, we investigated Flow Matching as a generative prior for high-resolution CT image reconstruction and provided a comprehensive comparison with diffusion-based approaches under sparse-view acquisition settings. By training a Rectified Flow Matching model on $256 \times 256$ chest CT images from the Mayo Clinic Low-Dose CT dataset, we demonstrated that Flow Matching can effectively learn a high-quality and anatomically meaningful prior, despite the limited size and variability typical of medical imaging datasets.

A key aspect of our approach is a two-stage training strategy that combines strong, anatomically informed data augmentation with a subsequent fine-tuning phase using reduced or no augmentation. The unconditional generation experiments and the corresponding KID evaluation show that this strategy improves sample fidelity and reduces augmentation-induced distortions, confirming that the learned flow captures both global anatomical structure and fine-grained details. These results provide a necessary sanity-check for the use of the model as a prior in inverse problems.

When employed for CT reconstruction, Flow Matching-based methods consistently outperform diffusion-based baselines across all angular configurations considered. In particular, FlowDPS achieves the best overall performance in terms of PSNR, SSIM, and LPIPS, indicating superior reconstruction accuracy and perceptual quality. Qualitative results further confirm that Flow Matching priors better preserve fine anatomical structures and reduce reconstruction artifacts, especially in severely undersampled regimes.

From a practical standpoint, our experiments highlight that successfully applying generative-prior posterior sampling methods to CT requires careful numerical stabilization. Addressing the mismatch between the normalized image domain and the physical projection domain, controlling the amplification induced by the CT operator, and mitigating the inexactness of backprojection as an adjoint are all crucial for stable and fair evaluation. The normalization, gradient scaling, and CGLS-based corrections adopted in this work proved essential to ensure robust convergence across methods and acquisition settings.

Overall, the results of this study suggest that Flow Matching provides a stable, efficient, and accurate alternative to diffusion models for CT reconstruction, particularly in high-resolution and sparse-view scenarios. Beyond the specific methods evaluated here, the deterministic and continuous-time nature of Flow Matching opens promising directions for future research, including adaptive or higher-order solver strategies, tighter integration with physics-based constraints, and extensions to three-dimensional and dynamic CT reconstruction. We believe that the public release of the trained model and code will facilitate further investigation of Flow Matching priors in medical imaging and contribute to the development of reproducible and robust generative reconstruction methods.

\bibliographystyle{plain}
\bibliography{biblio}

\appendix
\section*{Appendix}
\section{Hyperparameter setting and grid search}
\label{sec:supp_hyperparams}

The performance of posterior sampling methods for inverse problems depends critically on the choice of the guidance scale parameter, denoted as $\nu_t$. This parameter regulates the trade-off between the data-consistency term, which ensures fidelity to the acquired sinogram measurements, and the generative prior, which ensures the anatomical plausibility of the reconstruction. However, the physical meaning of this parameter is slightly different among each method. In the following, we briefly remark, for each FM-based method, how $\nu_t$ is used:

\begin{itemize}
    \item \textbf{FlowDPS:} The update rule of FlowDPS is 
    \begin{align}
        \x_{t + \Delta t}= \x_t + \Delta t \vtheta(\x_t, t) - \nu_t \cdot c_t \cdot  \nabla_{\x_t} || \K \hat{\x}_{1 : t} - \y ||_2^2,
    \end{align}
    where
    \begin{align}
        c_t := \frac{|| \vtheta(\x_t, t) ||_2}{|| \nabla_{\x_t} || \K \hat{\x}_{1 : t} - \y ||_2^2 ||_2}
    \end{align}
    is a normalization coefficient.  Here, $\nu_t$ effectively represents the balancing term between the prior and the likelihood distributions;
    \item \textbf{PnP-Flow:} The update rule for PnP-Flow is
    \begin{align}
        \begin{cases}
            &\hat{\x}_{t + \Delta t} = \x_t + \Delta t \vtheta(\x_t, t), \\
            &\x_{t + \Delta t} = \x_t - \nu_t \cdot  \nabla_{\x_t} || \K \hat{\x}_{1 : t} - \y ||_2^2.
        \end{cases}
    \end{align}
    Therefore, the physical meaning of $\nu_t$ for PnP-Flow is similar (but not identical) to FlowDPS. In particular, we expect $\nu_t$ to be smaller for PnP-Flow. 
    \item \textbf{FLOWER:}  The update rule for FLOWER is
    \begin{align}
    \begin{cases}
            &\hat{\x}_{t + \Delta t} = \x_t + \Delta t \vtheta(\x_t, t), \\
            &\x_{t + \Delta t} = \x_t - \nu_t \cdot \K^T (\K \hat{\x}_{1 : t} - \y).
    \end{cases}
    \end{align}
    which is similar to how $\nu_t$ is used for PnP-Flow;
    \item \textbf{ICTM:} the update rule for ICTM is
    \begin{align}
    \begin{cases}
        &\hat{\x}_{t + \Delta t} = \x_t + \Delta t \vtheta(\x_t, t), \\
        &\x_{t + \Delta t}^{proj} = \texttt{CGLS}(\arg\min_{\x} || \K \x - \y||_2^2, \x^{(0)} = \hat{\x}_{t + \Delta t}, \texttt{maxit}=2), \\
        &\x_{t + \Delta t} = (1 - \nu_t) \cdot \hat{\x}_{t + \Delta t} + \nu_t \cdot \x_{t + \Delta t}^{proj},
    \end{cases}
    \end{align}
    where $\texttt{CGLS}(\arg\min_{\x} || \K \x - \y||_2^2, \x^{(0)} = \hat{\x}_{t + \Delta t}, \texttt{maxit}=2)$ indicates the application of the CGLS algorithm for $2$ iterations using $\hat{\x}_{t + \Delta t}$ as a starting iterate. Here, the parameter $\nu_t$ is constrained to be in the range $[0, 1]$ and represents the balancing term between the estimated $\x_{t+\Delta t}$ by the FM-model and its projection over the solution of $|| \K\x - \y ||_2^2$.
\end{itemize}
To ensure a fair comparison between Flow Matching-based methods (FlowDPS, PnP-Flow, FLOWER, ICTM) and diffusion-based baselines (DPS, DDRM), we performed a systematic grid search on a hold-out validation set consisting of approximately $350$ slices from the Mayo Clinic dataset. For each method and each angular configuration ($60$, $90$, and $120$ degrees), we evaluated reconstruction performance over a range of candidate values, first by changing $\nu_t$ logarithmically, and then in a more fine-grained grid.

We selected the optimal $\nu_t$ that maximized the Peak Signal-to-Noise Ratio (PSNR) on the validation set. Then, we set the number of time stamps $N = \frac{1}{\Delta t}$ for each method individually to balance between the reconstruction time and quality. The final values used for the experiments are summarized as follows:
\begin{itemize}
    \item \textbf{FlowDPS:} $\nu_t = 5.0$ (60/90 angles), $\nu_t = 1.0$ (120 angles), $N = 500$ steps.
    \item \textbf{PnP-Flow:} $\nu_t = 0.4$ (60/90 angles), $\nu_t = 0.3$ (120 angles), $N = 200$ steps.
    \item \textbf{FLOWER:} $\nu_t = 50$ (60/90 angles), $\nu_t = 25$ (120 angles), $N = 300$ steps.
    \item \textbf{ICTM:} $\nu_t = 0.9$ (60/90 angles), $\nu_t = 0.8$ (120 angles), $N = 100$ steps.
\end{itemize}
It is worth noting that FlowDPS and FLOWER generally required larger scale magnitudes compared to PnP-Flow and ICTM, reflecting the different ways data consistency is injected into the ODE integration (gradient-based guidance vs. proximal/projection steps).
For comparison, we also report the value of the hyperparameters associated with DPS and DDRM used in the comparison:
\begin{itemize}
    \item \textbf{DPS:} $\zeta_t = 0.01$ (60 angles), $\zeta_t = 0.015$ (90 angles), $\zeta_t = 0.02$ (120 angles), $N = 100$ steps (60 angles), $N = 150$ steps (90/120 angles).
    \item \textbf{DDRM:} CGLS iterations $= 2$ (60/90/120 angles), $N = 50$ steps (60/90/120 angles).
\end{itemize}

\section{More experimental results}
\label{sec:supp_experiments}

To further validate the robustness of the proposed Flow Matching prior, we provide additional quantitative and qualitative results on specific individual samples from the test set. While the main manuscript reports metrics averaged over the entire test dataset, examining individual cases allows for a more granular analysis of how different methods handle specific anatomical features and varying noise realizations.

In this section, we focus on two representative slices, referred to as \texttt{C081-45} and \texttt{C081-79}. We evaluate the reconstruction quality under sparse-view settings with $60$, $90$, and $120$ projection angles.

\begin{table}[t]
\centering
\caption{Reconstruction results (PSNR $\uparrow$, SSIM $\uparrow$, LPIPS $\downarrow$) for samples C081-45 and C081-79 across different projection angles.}
\label{tab:recon_results_by_angles}

\setlength{\tabcolsep}{3pt} 
\begin{tabular}{cl ccc @{\hskip 0.5cm}ccc}
\toprule
& & \multicolumn{3}{c}{\texttt{C081-45}} & \multicolumn{3}{c}{\texttt{C081-79}} \\
\cmidrule(lr){3-5} \cmidrule(lr){6-8}
&\textbf{Method} & \textbf{PSNR} & \textbf{SSIM} & \textbf{LPIPS} & \textbf{PSNR} & \textbf{SSIM} & \textbf{LPIPS} \\
\midrule

\multirow{6}{*}{\rotatebox{90}{\textbf{Angles = $60$}}}
&DPS       & $28.36$ & $0.7956$ & $0.3399$  & $28.13$ & $0.7724$ & $0.3413$ \\
&DDRM      & $35.63$ & $0.8943$ & $0.2660$  & $34.57$ & $0.8571$ & $0.2765$ \\
\cdashlinelr{2-8} 
&FlowDPS   & $\mathbf{36.50}$ & $\mathbf{0.9182}$ & $\mathbf{0.1766}$ & $\mathbf{35.28}$ & $\mathbf{0.9043}$ & $\mathbf{0.2037}$ \\
&PnP-Flow  & $34.17$ & $0.8838$ & $0.2450$  & $33.48$ & $0.8739$ & $0.2607$ \\
&FLOWER    & $34.74$ & $0.8382$ & $0.2724$  & $34.39$ & $0.8558$ & $0.2415$ \\
&ICTM      & $34.67$ & $0.8666$ & $0.2696$  & $34.41$ & $0.8749$ & $0.2570$ \\
\midrule
\midrule

\multirow{6}{*}{\rotatebox{90}{\textbf{Angles = $90$}}}
&DPS       & $30.58$ & $0.8261$ & $0.3502$  & $30.39$ & $0.8259$ & $0.3307$ \\
&DDRM      & $36.14$ & $0.9131$ & $0.2548$  & $35.34$ & $0.9025$ & $0.2544$ \\
\cdashlinelr{2-8}
&FlowDPS   & $\mathbf{36.84}$ & $\mathbf{0.9238}$ & $\mathbf{0.1679}$ & $35.58$ & $\mathbf{0.9103}$ & $\mathbf{0.2026}$ \\
&PnP-Flow  & $34.99$ & $0.8953$ & $0.2331$  & $34.20$ & $0.8873$ & $0.2296$ \\
&FLOWER    & $36.64$ & $0.9116$ & $0.2391$  & $\mathbf{35.92}$ & $0.9033$ & $0.2467$ \\
&ICTM      & $34.86$ & $0.8775$ & $0.2467$  & $34.67$ & $0.8780$ & $0.2422$ \\
\midrule
\midrule

\multirow{6}{*}{\rotatebox{90}{\textbf{Angles = $120$}}}
&DPS       & $31.42$ & $0.8628$ & $0.2928$  & $31.34$ & $0.8336$ & $0.2864$ \\
&DDRM      & $36.40$ & $0.9220$ & $0.2476$  & $\mathbf{35.46}$ & $\mathbf{0.9129}$ & $0.2539$ \\
\cdashlinelr{2-8}
&FlowDPS   & $\mathbf{36.83}$ & $\mathbf{0.9261}$ & $\mathbf{0.1571}$ & $35.37$ & $0.9103$ & $0.2141$ \\
&PnP-Flow  & $35.48$ & $0.9070$ & $0.2130$  & $34.61$ & $0.8981$ & $0.2269$ \\
&FLOWER    & $35.92$ & $0.9075$ & $0.1840$  & $34.98$ & $0.8991$ & $\mathbf{0.1853}$ \\
&ICTM      & $35.53$ & $0.8989$ & $0.2053$  & $35.05$ & $0.8957$ & $0.1915$ \\
\bottomrule
\end{tabular}
\end{table}

\subsection{Sample-Specific Quantitative and Qualitative Analysis}

We jointly analyze quantitative metrics and visual reconstructions for two representative test samples, \texttt{C081-45} and \texttt{C081-79}, under sparse-view acquisition with $60$, $90$, and $120$ projection angles. This combined analysis allows us to assess not only average performance trends, as reported in the main manuscript, but also how different reconstruction strategies behave on individual anatomical realizations.

Table~\ref{tab:recon_results_by_angles} reports PSNR, SSIM, and LPIPS values for both samples. Overall, the quantitative trends closely mirror those observed in the main paper. Flow Matching-based approaches consistently outperform diffusion baselines across all sparsity levels, with particularly large improvements in perceptual quality as measured by LPIPS. Among all methods, \textbf{FlowDPS} achieves the best overall performance on both samples in the most challenging sparse-view regime. For instance, at $60$ angles on sample \texttt{C081-45}, FlowDPS reaches $36.50$~dB PSNR and $0.9182$ SSIM, compared to $35.63$~dB and $0.8943$ for DDRM, while reducing LPIPS from $0.2660$ (DDRM) to $0.1766$. Similar gains are observed for sample \texttt{C081-79}, where FlowDPS yields the lowest LPIPS ($0.2037$ at $60$ angles) and the highest reconstruction fidelity among all methods.

\begin{figure}[tbhp!]
    \centering
    \setlength{\tabcolsep}{2pt}
    \resizebox{0.9\linewidth}{!}{
    \begin{tabular}{cc cccccc}
    & \Large{True} & \Large{DPS} & \Large{DDRM} & \Large{FlowDPS} & \Large{PnP-Flow} & \Large{FLOWER} & \Large{ICTM} \\
    \rotatebox{90}{\hspace{15px}\Large\textbf{Angles = $60$}}&
    \imgwithzoom{width=0.2\linewidth}{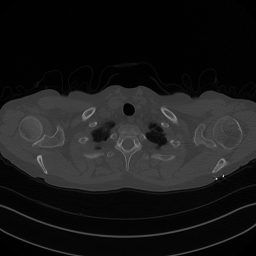}{0.5}{0.4}{0.2}{0.3}{1}{}&
    \imgwithzoom{width=0.2\linewidth}{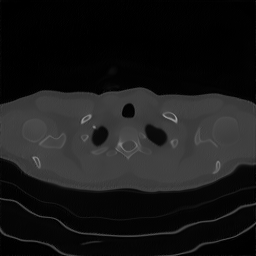}{0.5}{0.4}{0.2}{0.3}{0}{}&
    \imgwithzoom{width=0.2\linewidth}{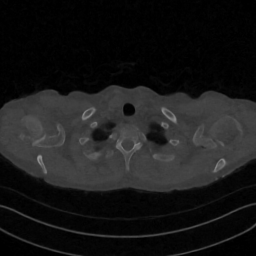}{0.5}{0.4}{0.2}{0.3}{0}{}&
    \imgwithzoom{width=0.2\linewidth}{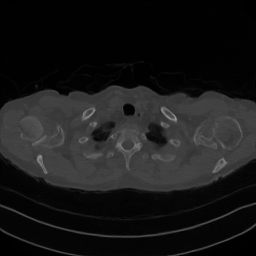}{0.5}{0.4}{0.2}{0.3}{0}{}&
    \imgwithzoom{width=0.2\linewidth}{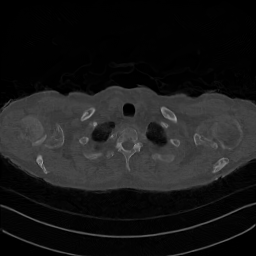}{0.5}{0.4}{0.2}{0.3}{0}{}&
    \imgwithzoom{width=0.2\linewidth}{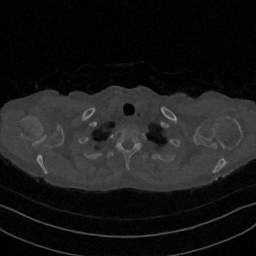}{0.5}{0.4}{0.2}{0.3}{0}{}&
    \imgwithzoom{width=0.2\linewidth}{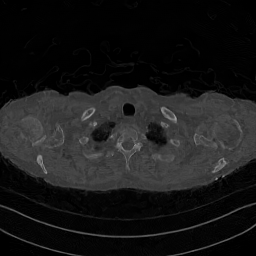}{0.5}{0.4}{0.2}{0.3}{0}{} \\
    
    \rotatebox{90}{\hspace{15px}\Large\textbf{Angles = $90$}}&
    \imgwithzoom{width=0.2\linewidth}{imgs/C081_45/C081_45.png}{0.5}{0.4}{0.2}{0.3}{1}{}&
    \imgwithzoom{width=0.2\linewidth}{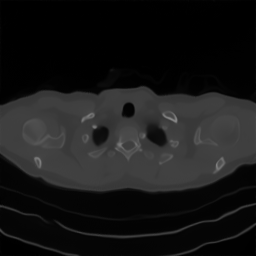}{0.5}{0.4}{0.2}{0.3}{0}{}&
    \imgwithzoom{width=0.2\linewidth}{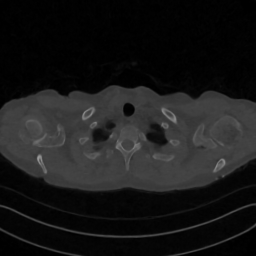}{0.5}{0.4}{0.2}{0.3}{0}{}&
    \imgwithzoom{width=0.2\linewidth}{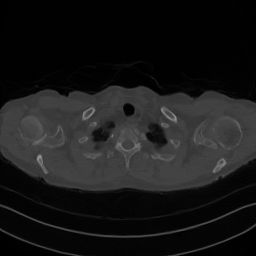}{0.5}{0.4}{0.2}{0.3}{0}{}&
    \imgwithzoom{width=0.2\linewidth}{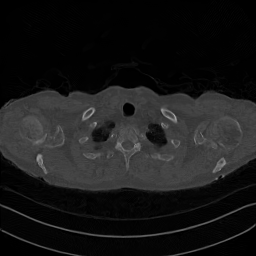}{0.5}{0.4}{0.2}{0.3}{0}{}&
    \imgwithzoom{width=0.2\linewidth}{imgs/C081_45/recon_flowers_st300_a60_s50.0/recon.png}{0.5}{0.4}{0.2}{0.3}{0}{}&
    \imgwithzoom{width=0.2\linewidth}{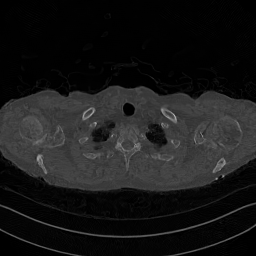}{0.5}{0.4}{0.2}{0.3}{0}{} \\

    \rotatebox{90}{\hspace{15px}\Large\textbf{Angles = $120$}}&
    \imgwithzoom{width=0.2\linewidth}{imgs/C081_45/C081_45.png}{0.5}{0.4}{0.2}{0.3}{1}{}&
    \imgwithzoom{width=0.2\linewidth}{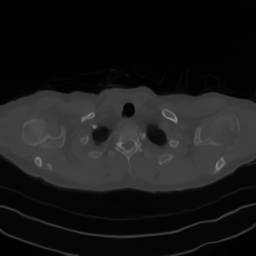}{0.5}{0.4}{0.2}{0.3}{0}{}&
    \imgwithzoom{width=0.2\linewidth}{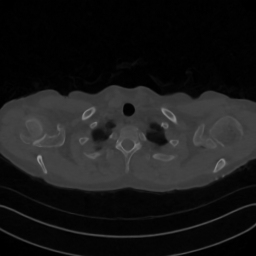}{0.5}{0.4}{0.2}{0.3}{0}{}&
    \imgwithzoom{width=0.2\linewidth}{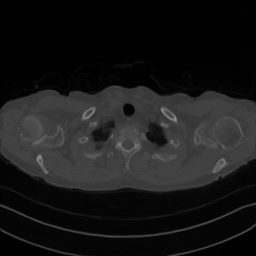}{0.5}{0.4}{0.2}{0.3}{0}{}&
    \imgwithzoom{width=0.2\linewidth}{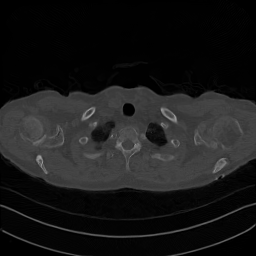}{0.5}{0.4}{0.2}{0.3}{0}{}&
    \imgwithzoom{width=0.2\linewidth}{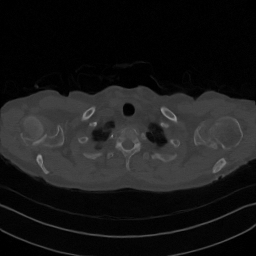}{0.5}{0.4}{0.2}{0.3}{0}{}&
    \imgwithzoom{width=0.2\linewidth}{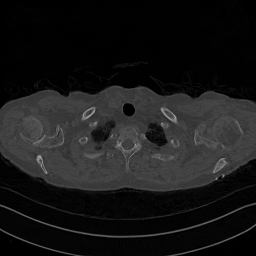}{0.5}{0.4}{0.2}{0.3}{0}{} \\
    \end{tabular}}
    \caption{Visual comparison for \textbf{Sample C081-45} across three sparse-view configurations. The red box indicates the zoomed region shown in the inset.}
    \label{fig:sample_45}
\end{figure}

\begin{figure}[tbhp!]
    \centering
    \setlength{\tabcolsep}{2pt}
    \resizebox{0.9\linewidth}{!}{
    \begin{tabular}{cc cccccc}
    & \Large{True} & \Large{DPS} & \Large{DDRM} & \Large{FlowDPS} & \Large{PnP-Flow} & \Large{FLOWER} & \Large{ICTM} \\
    \rotatebox{90}{\hspace{15px}\Large\textbf{Angles = $60$}}&
    \imgwithzoom{width=0.2\linewidth}{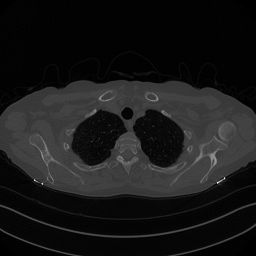}{0.5}{0.4}{0.2}{0.3}{1}{}&
    \imgwithzoom{width=0.2\linewidth}{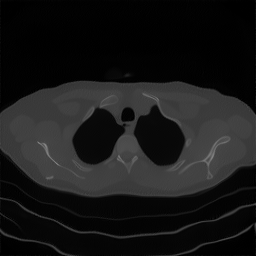}{0.5}{0.4}{0.2}{0.3}{0}{}&
    \imgwithzoom{width=0.2\linewidth}{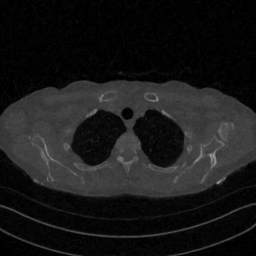}{0.5}{0.4}{0.2}{0.3}{0}{}&
    \imgwithzoom{width=0.2\linewidth}{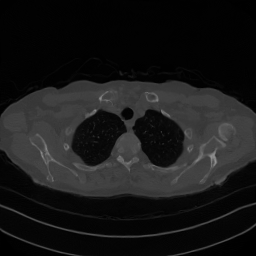}{0.5}{0.4}{0.2}{0.3}{0}{}&
    \imgwithzoom{width=0.2\linewidth}{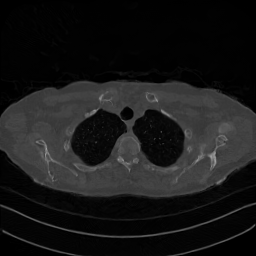}{0.5}{0.4}{0.2}{0.3}{0}{}&
    \imgwithzoom{width=0.2\linewidth}{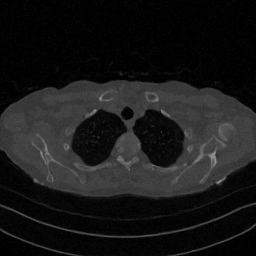}{0.5}{0.4}{0.2}{0.3}{0}{}&
    \imgwithzoom{width=0.2\linewidth}{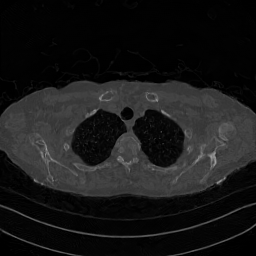}{0.5}{0.4}{0.2}{0.3}{0}{} \\
    
    \rotatebox{90}{\hspace{15px}\Large\textbf{Angles = $90$}}&
    \imgwithzoom{width=0.2\linewidth}{imgs/C081_79/C081_79.png}{0.5}{0.4}{0.2}{0.3}{1}{}&
    \imgwithzoom{width=0.2\linewidth}{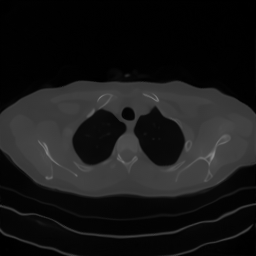}{0.5}{0.4}{0.2}{0.3}{0}{}&
    \imgwithzoom{width=0.2\linewidth}{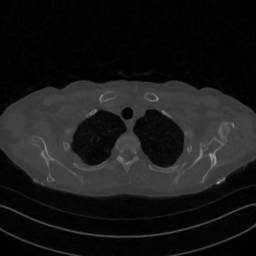}{0.5}{0.4}{0.2}{0.3}{0}{}&
    \imgwithzoom{width=0.2\linewidth}{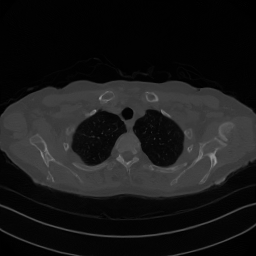}{0.5}{0.4}{0.2}{0.3}{0}{}&
    \imgwithzoom{width=0.2\linewidth}{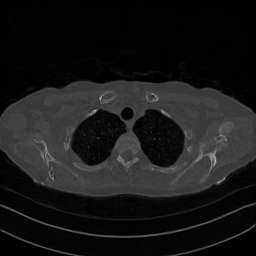}{0.5}{0.4}{0.2}{0.3}{0}{}&
    \imgwithzoom{width=0.2\linewidth}{imgs/C081_79/recon_flowers_st300_a60_s50.0/recon.png}{0.5}{0.4}{0.2}{0.3}{0}{}&
    \imgwithzoom{width=0.2\linewidth}{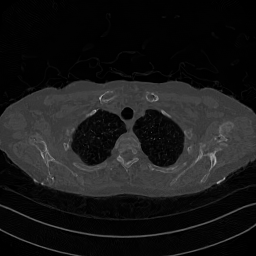}{0.5}{0.4}{0.2}{0.3}{0}{} \\

    \rotatebox{90}{\hspace{15px}\Large\textbf{Angles = $120$}}&
    \imgwithzoom{width=0.2\linewidth}{imgs/C081_79/C081_79.png}{0.5}{0.4}{0.2}{0.3}{1}{}&
    \imgwithzoom{width=0.2\linewidth}{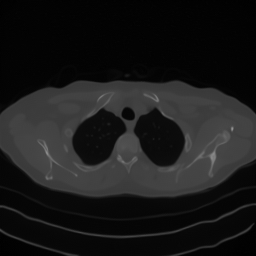}{0.5}{0.4}{0.2}{0.3}{0}{}&
    \imgwithzoom{width=0.2\linewidth}{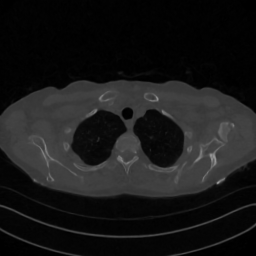}{0.5}{0.4}{0.2}{0.3}{0}{}&
    \imgwithzoom{width=0.2\linewidth}{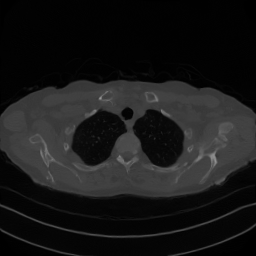}{0.5}{0.4}{0.2}{0.3}{0}{}&
    \imgwithzoom{width=0.2\linewidth}{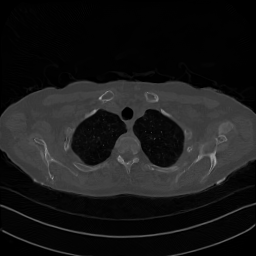}{0.5}{0.4}{0.2}{0.3}{0}{}&
    \imgwithzoom{width=0.2\linewidth}{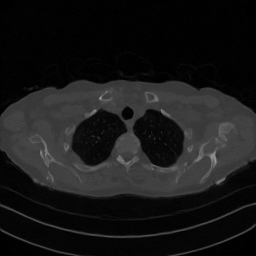}{0.5}{0.4}{0.2}{0.3}{0}{}&
    \imgwithzoom{width=0.2\linewidth}{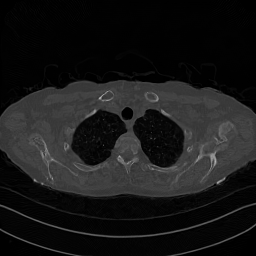}{0.5}{0.4}{0.2}{0.3}{0}{} \\
    \end{tabular}}
    \caption{Visual comparison for \textbf{Sample C081-79} across three sparse-view configurations. The red box indicates the zoomed region shown in the inset.}
    \label{fig:sample_79}
\end{figure}

These quantitative improvements are reflected in the visual reconstructions shown in Figure~\ref{fig:sample_45} and Figure~\ref{fig:sample_79}. For sample \texttt{C081-45} (Figure~\ref{fig:sample_45}), diffusion-based methods recover the global anatomical structure but suffer from noticeable over-smoothing under severe sparsity. DPS produces washed-out reconstructions across all angular settings, while DDRM improves structural coherence but introduces slight blurring of fine anatomical details, as highlighted in the zoomed regions. In contrast, Flow Matching-based methods yield sharper reconstructions with improved preservation of small structures. FlowDPS in particular effectively suppresses streaking artifacts while maintaining sharp bone boundaries and fine anatomical texture. PnP-Flow and ICTM generate similarly sharp reconstructions but introduce a small amount of residual noise, consistent with their slightly inferior LPIPS scores. FLOWER produces visually convincing results but fails to recover some of the smallest structures visible in the zoomed regions.

Sample \texttt{C081-79} (Figure~\ref{fig:sample_79}) represents a more challenging reconstruction case, as also evidenced by the uniformly lower quantitative metrics across all methods in Table~\ref{tab:recon_results_by_angles}. Under $60$-view acquisition, diffusion-based methods again exhibit pronounced smoothing and loss of fine details, whereas Flow Matching-based approaches better preserve anatomical features and reduce visually salient artifacts. FlowDPS remains the most robust method in this severely underdetermined regime, although the performance margin relative to other FM-based methods is reduced compared to sample \texttt{C081-45}.

As the number of projection angles increases, all methods benefit from the improved conditioning of the inverse problem. At $120$ angles, DDRM becomes competitive with FlowDPS on sample \texttt{C081-79}, slightly surpassing it in PSNR and SSIM. However, the differences remain small, and FlowDPS continues to achieve comparable visual quality with reduced perceptual artifacts. This behavior suggests that while diffusion priors can perform well in relatively well-conditioned settings, Flow Matching priors retain a clear advantage in sparse-view regimes, offering improved robustness and perceptual fidelity.

Overall, this combined quantitative and qualitative analysis confirms that the conclusions drawn in the main manuscript generalize to individual test samples. Flow Matching-based methods consistently provide superior anatomical coherence and perceptual quality in challenging sparse-view settings, while diffusion-based approaches become competitive only as the acquisition geometry becomes less ill-posed.

\end{document}